\documentclass{article}

\usepackage{spconf}
\usepackage{times}
\usepackage{epsfig}
\usepackage{graphicx}
\usepackage{amsmath}
\usepackage{amssymb}
\usepackage{color}
\usepackage{xspace}
\graphicspath{{./img/}}

\definecolor{gg}{RGB}{0,100,0}

\newcommand{\soa}{state-of-the-art\xspace}

\usepackage[pagebackref=true,breaklinks=true,colorlinks,bookmarks=false]{hyperref}

\begin{document}

%%%%%%%%% TITLE
\title{Texture-Aware Superpixel Segmentation}

% \author{Rémi Giraud\\
% Univ. Bordeaux, IMS, CNRS, UMR 5218,\\
% F-33400 Talence, France.\\
% Bordeaux INP, LaBRI, UMR 5800, \\
% F-33405 Talence, France.\\
% {\tt\small remi.giraud@u-bordeaux.fr}
% \and

\name{R{\'e}mi Giraud$^{1}$ 
\qquad Vinh-Thong Ta$^{2}$ \qquad Nicolas Papadakis$^{3}$
\qquad Yannick Berthoumieu$^{1}$
 % \thanks{
%   \vspace{-0.05cm}
%  \hspace{-0.75cm} 
% Preprint available at: {\url{https://arxiv.org/abs/1901.11111}}
%   \vspace{-0.05cm}
  %}
}
\address{$^{1}$Bordeaux INP, Univ. Bordeaux, CNRS, IMS,  UMR 5218, F-33400 Talence, France.\\
%     $^{2}$Bordeaux INP, IMS, UMR 5218, F-33400 Talence, France.\\
    $^{2}$Bordeaux INP, Univ. Bordeaux,  CNRS, LaBRI, UMR 5800, F-33400 Talence, France.\\
    $^{3}$CNRS, Univ. Bordeaux, IMB, UMR 5251, F-33400 Talence, France.\\
}

\maketitle
%\thispagestyle{empty}

%%%%%%%%% ABSTRACT
\begin{abstract}

Most superpixel algorithms
compute a trade-off between spatial and color features at the pixel level.
Hence, they may need fine parameter tuning to balance the two measures,
and highly fail
to group pixels with similar local texture properties.
In this paper, we address these issues with a new Texture-Aware SuperPixel (TASP) method.
To accurately segment textured and smooth areas,
TASP automatically adjusts its spatial constraint according to the local feature variance.
Then, to ensure texture homogeneity within superpixels, 
a new pixel to superpixel patch-based distance is proposed.
TASP outperforms the segmentation accuracy of the \soa methods 
 on texture and also natural color image datasets.
\end{abstract} 
\vspace{0.2cm}

\noindent
 \vspace{-0.165cm}

%%%%%%%%% BODY TEXT
\section{Introduction}
\label{sec:intro}

Superpixel segmentation approaches that locally group pixels into regions have become very popular in image processing and computer vision applications.
The aim is to exploit the local redundancy of information
to lower the computational burden 
and to potentially improve the performances by reducing the noise of a processing at the pixel level.
Superpixels can also be considered as a multi-resolution approach
that preserves image contours, contrary to standard regular  downsampling methods.
It is thus a very interesting pre-processing for applications such as 
visual saliency estimation \cite{liu2014superpixel,he2015supercnn},
data association across views \cite{sawhney2014},
segmentation and classification  \cite{gould2014,gadde2016superpixel,giraud2017_spm} and 
object detection \cite{arbelaez2011,shu2013improving,yan2015object} %
or tracking \cite{chang2013,reso2013}.

For the past years, most superpixel methods have tended to produce equally-sized regions with homogeneous pixels in terms of color.
This paradigm is usually in line with the segmentation of a natural image objects, whose contours can be detected by color changes.
Hence, to cluster the pixels into regions, state-of-the-art methods such as 
\cite{liu2011,vandenbergh2012,achanta2012,buyssens2014,achanta2017superpixels},
only use distance terms in  spatial and color (\emph{e.g.}, CIELab) spaces.
In \cite{liu2016manifold,chen2017}, more advanced feature spaces are defined to improve the segmentation performances.
More recently, \cite{giraud2018_scalp} proposes to consider contour information in the clustering to ensure the respect of the object boundaries.
Nevertheless, such framework requires the need for prior contour detection, at the expense of a global higher
complexity.

Most of these state-of-the-art methods only use the pixel information as clustering feature.
Therefore, they can be severely impacted by high frequency contrast variations and fail to 
produce equally-sized regions
having the same textural properties.
The proposed method TASP is compared in Figure \ref{fig:intro} to the state-of-the-art approaches on a synthetic texture image.
While TASP produces a relevant segmentation,
all other methods
are highly misleaded by the texture patterns.
We used the  regularity parameters recommended by the authors for tunable methods \cite{achanta2012,buyssens2014,yao2015,chen2017,giraud2018_scalp},
but no other setting would enable to capture texture information.
Superpixel methods are indeed generally optimized and evaluated on noise-free natural color images, although
specific tasks require to decompose
highly textured or low resolution grayscale images,
for instance in medical applications \cite{tian2016superpixel}.

\newcommand{\ww}{0.19\textwidth}
\newcommand{\hh}{0.111\textwidth}
\begin{figure*}[ht] 
\centering 
{\footnotesize
\centering
\renewcommand{\arraystretch}{0.95}
\begin{tabular}{@{\hspace{0mm}}c@{\hspace{1mm}}c@{\hspace{1mm}}c@{\hspace{1mm}}c@{\hspace{1mm}}c@{\hspace{0mm}}}
%  & $|\mathcal{S}|=80$& $|\mathcal{S}|=80$& $|\mathcal{S}|=80$\\
  \includegraphics[width=\ww,height=\hh]{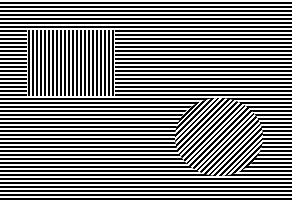}& 
  \includegraphics[width=\ww,height=\hh]{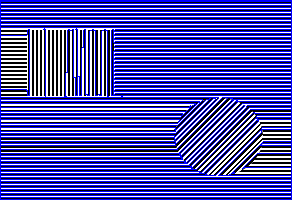}&
  \includegraphics[width=\ww,height=\hh]{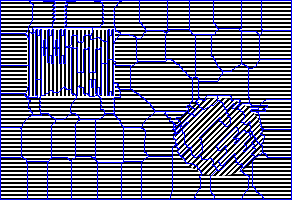}&                                                                                                                                                                                                                                                                                    
  \includegraphics[width=\ww,height=\hh]{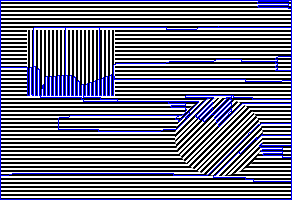}&    
  \includegraphics[width=\ww,height=\hh]{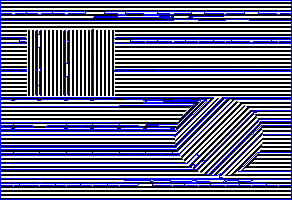}\\
  Initial image & ERS \cite{liu2011} %- ($|\mathcal{S}|=80$)
   & {\color{black}SEEDS} \cite{vandenbergh2012} & SLIC \cite{achanta2012}  & ERGC \cite{buyssens2014}   \\
  \includegraphics[width=\ww,height=\hh]{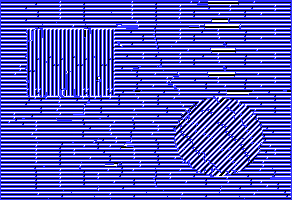}& %2015 
  \includegraphics[width=\ww,height=\hh]{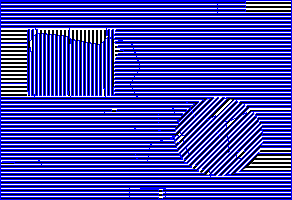}& %2017
  \includegraphics[width=\ww,height=\hh]{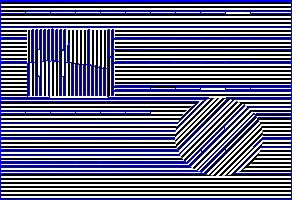}& %2017
  \includegraphics[width=\ww,height=\hh]{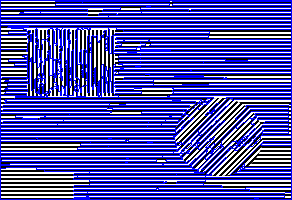}& %2018
  \includegraphics[width=\ww,height=\hh]{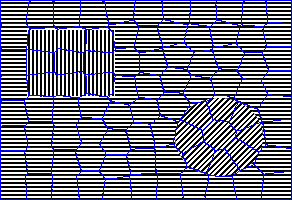}\\
ETPS \cite{yao2015} & LSC \cite{chen2017}    & SNIC \cite{achanta2017superpixels}   &  SCALP \cite{giraud2018_scalp}  & TASP  \\[-1.75ex]
 \end{tabular}
 }
\caption{
Comparison of the TASP method to state-of-the-art approaches on a synthetic texture image for $100$ superpixels.
Only TASP succeeds in capturing the textures while other methods are highly misleaded by high frequency contrast variations.
} \vspace{-0.3cm}
\label{fig:intro} 
\end{figure*}

To overcome the limitations of handcrafted color spaces, 
deep learning approaches have been proposed \cite{jampani2018superpixel,liu2018learning}.
Nevertheless, the gain obtained with learned features on a training dataset 
may come at the expense of usual deep learning limitations, 
\emph{i.e.}, important learning time,
need for a substantial training database and material resources,
and direct applicability limited to similar images.
Moreover, 
these approaches do not directly extend to supervoxels and 
prevent from setting the shape regularity, 
which can highly impact the performances of superpixel-based tasks.
It is thus still necessary to increase the robustness of
non-deep  learning
superpixel methods  to textures, while preserving their desired properties:  
adaptability, low complexity and limited parameter settings.

  \vspace{0.035cm}

\textbf{Contributions.} 
In this work, we propose a new Texture-Aware SuperPixel (TASP) clustering method 
able to accurately segment highly textured images,
but also any input image,
\emph{e.g.}, natural color ones,
using the same parameters.

To be able to generate relevant superpixels
on textured images (see Figure \ref{fig:intro}),
TASP adjusts its spatial constraint, according to the feature variance within the superpixel.
This way, TASP also addresses the need for fine manual regularity setting.
Most recent state-of-the-art methods globally set this parameter according to the image nature, leading
default or sub-optimal settings to highly impact the results 
\cite{giraud2017_gef}.

Then, to ensure the texture homogeneity,
we introduce a new patch-based framework  
enabling to easily evaluate the similarity of
a pixel neighborhood to a superpixel.

We validate TASP
on natural color images  from a standard segmentation dataset \cite{martin2001}, 
and on two new datasets proposed to evaluate texture segmentation performances.
TASP significantly outperforms  the \soa methods
on texture segmentation performances, while performing as well, or better, on natural images,
using the same parameters. 

\vspace{-0.035cm}

\section{Texture-Aware SuperPixels}

The TASP method improves the superpixel decomposition approach used in \cite{achanta2012,chen2017,giraud2018_scalp}, that is first presented in this section.
Then, we propose a method to locally set the spatial regularity of superpixels,  to automatically adapt to the image content. 
Finally, we introduce a new pixel to superpixel texture homogeneity measure
to group pixels in terms of texture.

\vspace{-0.2cm}

\subsection{K-means-based Iterative Clustering}

The standard framework of \cite{achanta2012} only requires the number of superpixels to produce and a regularity parameter.
The algorithm is based on an iteratively constrained K-means clustering of pixels.
Superpixels $S_i$ are first regularly set over the image domain as blocks of size $s{\times}s$,
and are described by their average intensity feature $F_{S_i}$ (CIELab colors for \cite{achanta2012})
and their spatial barycenter $X_{S_i}=[x_i,y_i]$ of pixels in $S_i$.
The clustering relies on 
a feature  $d_F(F_p,F_{S_i})$=${\|F_p-F_{S_i}\|}_2$, 
and a spatial distance term $d_s(X_p,X_{S_i})$=${\|X_p-X_{S_i}\|}_2$.
At each iteration,
each superpixel $S_i$ is compared to all pixels $p$, of feature $F_p$ at position $X_p$, within a $(2s$$+$$1)$${\times}$$(2s$$+$$1)$ region around its barycenter $X_{S_i}$.
A pixel $p$ is associated to the superpixel $S_i$ minimizing the distance $D$ defined as: \\[-1.75ex]
{\small
\begin{equation}
D(p,S_i)=d_F(F_p,F_{S_i}) + d_s(X_p,X_{S_i})\frac{m^2}{s^2} ,  \label{distance_slic} 
\end{equation}
}%
with $m$ the parameter setting the superpixel shape regularity. 
A post-processing step finally ensures region connectivity.

Although this method can accurately gather pixels having similar colors, 
 $m$ is globally set and cannot adapt to all  local image contours. It also highly fails to capture texture patterns, 
as it only considers feature information at the pixel level.

\subsection{Local Adaptation of Superpixel Regularity}

For most methods, including \cite{achanta2012,chen2017,giraud2018_scalp},
the regularity parameter $m$ must be manually set, according to the dynamic of the feature term $d_F$.
Hence, default parameters for natural color images may lead model \eqref{distance_slic} to generate highly irregular clustering on textures, and the post-processing step enforcing connectivity to irrelevantly merge regions (see Figure \ref{fig:intro}).
We address this issue
by using for each superpixel $S_i$,
a regularity parameter $m_i$ defined
 according to 
 the feature variance  $\sigma_i(F_p)$ of all pixels $p\in S_i$ such that:
\\[-1ex]
{\small
\begin{equation}
 m_{i} = {m}  \exp{\left(\frac{\sigma_i(F_{p})^2}{\beta}\right)}  ,
 \label{laser}
 \end{equation} \\[-1.5ex]
}%
with a scaling parameter $\beta$. 
Such regularity term $m_i$
is able to increase 
the spatial constraint
in the TASP model \eqref{tasp} for superpixels having high feature variances, and
to reduce it in smooth areas, 
so the superpixel boundaries can capture image objects that are perceptible from limited feature variations.

\begin{figure}[t!]
\newcommand{\kkk}{0.145\textwidth}
\newcommand{\kk}{0.095\textwidth}
\begin{center}
{\footnotesize
 \begin{tabular}{@{\hspace{0mm}}c@{\hspace{1mm}}c@{\hspace{1mm}}c@{\hspace{1mm}}c@{\hspace{0mm}}}  \includegraphics[width=0.115\textwidth,height=\kkk]{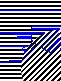}&
  \includegraphics[width=0.115\textwidth,height=\kkk]{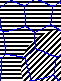}&
  \includegraphics[width=0.115\textwidth,height=\kkk]{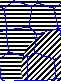}&
  \includegraphics[width=0.115\textwidth,height=\kkk]{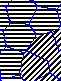}\\
  \includegraphics[width=0.115\textwidth,height=\kk]{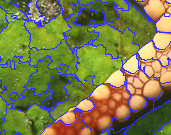}&
  \includegraphics[width=0.115\textwidth,height=\kk]{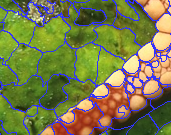}&
  \includegraphics[width=0.115\textwidth,height=\kk]{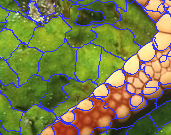}&
  \includegraphics[width=0.115\textwidth,height=\kk]{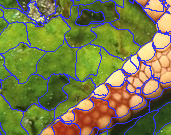}\\
 (a) SLIC \cite{achanta2012} & (b) w/ \eqref{laser} & (c) w/ \eqref{laser}, \eqref{pbd} & (d) w/ \eqref{laser}, \eqref{pbd}, \eqref{unicity} \\[-3.5ex]
 \end{tabular}
 }
 \end{center}
\caption{
SLIC  \cite{achanta2012} with optimal regularity for color images (a)
vs TASP contributions (b)-(d), that accurately decomposes both texture and color images with the same parameters (d).
} \vspace{-0.335cm}
\label{fig:unicity} 
\end{figure}

This way, without manually adapting $m$ in \eqref{laser},
TASP can compute relevant superpixels on both
highly textured images
(see Figure \ref{fig:unicity}(b)-top),
and natural color ones
(see Figure \ref{fig:unicity}(b)-bottom), 
since $m_i$
\eqref{laser} automatically adjusts the trade-off between $d_F$ and $d_s$ in \eqref{tasp}.
Nevertheless, the clustering accuracy  still has to be 
improved to capture texture information.

\subsection{Texture Homogeneity Measure}

\subsubsection{Pixel to Superpixel Patch-based Distance}

In this section, we propose a method to measure the texture similarity between a pixel neighborhood
and the content of a superpixel,
thus between two regions of different sizes.
A texture descriptor at patch and superpixel levels would yield higher complexity
and additional parameter settings.
Moreover, 
texture cannot be preserved
as well as for color and spatial information
with a global average over the whole superpixel.
The framework must preserve its limited complexity, 
and to be able to adapt to any image content without any prior information.
Such constraints also prevent from using costly dictionary or learning-based approaches.

\begin{figure}[t!] 
\begin{center}
 \begin{tabular}{@{\hspace{1mm}}c}
  \includegraphics[width=0.45\textwidth,height=0.185\textwidth]{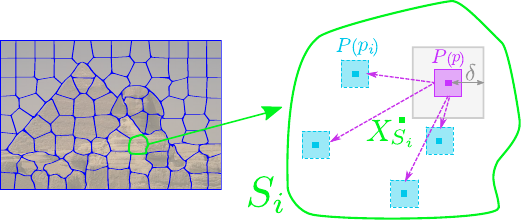}\\[-3.5ex]
 \end{tabular} 
 \end{center}
\caption{Selection of similar patches $P(p_i)$ in a superpixel $S_i$ of barycenter $X_{S_i}$,
outside a $\delta$-neighborhood, to
compute the texture homogeneity term \eqref{pbd} for a patch $P(p)$.
} \vspace{-0.35cm}
\label{fig:tasp} 
\end{figure}

To address these issues, 
we propose a new framework
using square patches 
to naturally capture texture information.
For a pixel $p$, of patch $P(p)$, and a superpixel $S_i$,
a nearest neighbor algorithm (see section \ref{sec:pm}) is used to find similar patches $P(p_i)$
such that $p_i\in S_i$, and
outside a $\delta$-neighborhood around $p$ (see Figure \ref{fig:tasp}).
The new term $d_P$ computes the average distance 
to the selected $p_i\in S_i$:
\\[-1.5ex]
{\small
\begin{equation}
 d_P(p,S_i) = \frac{1}{N}\sum_{p_i\in \mathcal{K}_{p}}  d_P(p,p_i) , \label{pbd}  \\[-0.5ex]
\end{equation}
}%
with $\mathcal{K}_{p}$ the set of $N$ selected pixels $p_i\in S_i$,
compared with a patch distance in the feature space,
such that 
  $d_P(p,p_i)=\frac{1}{n}{\|F_{P(p)}-F_{P(p_i)})\|}_2$, 
with $n$ the patch size. 

Any feature can be used in term \eqref{pbd}.
This way, we propose a general model that can easily evaluate the texture compliance 
of a pixel neighborhood to a superpixel, % in terms of texture,
while leveraging the need for complex texture classification approaches.

\vspace{-0.125cm}

\subsubsection{\label{sec:pm}Patch-based Nearest Neighbor Search}
The search for similar patches can be performed by any nearest neighbor (NN) method.
We choose to use PatchMatch, a fast iterative approximate-NN algorithm based on the propagation of good matches from adjacent neighbors  \cite{barnes2009}.
The computation of $d_P(p,S_i)$ can be directly performed for all pixels $p$ in the $(2s+1){\times}(2s+1)$ area around the barycenter $X_{S_i}$ of $S_i$.
The algorithm being partly random, $N$ patches in $S_i$ can be selected in parallel for each pixel $p$, to increase the robustness of the texture homogeneity term \eqref{pbd}. 

 \vspace{-0.05cm}

\subsubsection{Texture Unicity within Superpixels}

In the texture term \eqref{pbd}, the patch similarity is computed regardless of any spatial information.
Hence, a pixel $p$ to cluster may find similar local textures in restricted areas, 
leading a superpixel to potentially group 
several textures (Figure \ref{fig:unicity}(c)).

To ensure the texture unicity within a superpixel $S_i$, 
we consider in $d_P$ the spatial distance between the selected patches $P(p_i)$, at position $X_{p_i}\in S_i$, and $X_{S_i}$, the spatial barycenter of the superpixel $S_i$ such that:\\[-1.5ex]
{\small
\begin{equation}
d_P(p,p_i) = \frac{1}{n}{\|F_{P(p)}-F_{P(p_i)})\|}_2  + \frac{m_{i}^2}{s^2}\hspace{0.05cm} \Gamma \big(X_{p_i},X_{S_i}\big) ,  \label{unicity}
\end{equation}
}%
with $\Gamma$, a scaling function defined such that  $\Gamma(X_{p_i},X_{S_i}) = 2s^2(1-\exp{(-{\|X_{p_i}-X_{S_i}\|}_2^2/s^2)})$.
Such term iteratively contributes to restrict the search area and the diversity of textures within $S_i$
by highly penalizing similar patches found far from $X_{S_i}$.
Hence, the barycenter is encouraged to move to a homogeneous textured area 
and to be contained within the superpixel (see Figure \ref{fig:unicity}(d)).
This also increases the shape regularity, which is a desirable property
\cite{giraud2017_gef}.

\smallskip

The clustering distance in TASP is finally computed as: \\[-1.5ex]
{\small
\begin{equation}
D(p,S_i) =  d_F(F_{p},F_{S_i}) + d_s(X_{p},X_{S_i})\frac{m_{i}^2}{s^2} + {d_P(p,S_i)} . \label{tasp}
\end{equation} 
}%
This way, TASP becomes a very general method, 
able to handle textures,
and efficient on various image types with the exact same parameters, 
as demonstrated in the next section.

\vspace{-0.15cm}

\begin{figure*}[ht]
\renewcommand{\arraystretch}{0.95}
\renewcommand{\ww}{0.185\textwidth}
\renewcommand{\hh}{0.1225\textwidth}
\centering
{\color{black}
{\footnotesize
 \begin{tabular}{@{\hspace{0mm}}c@{\hspace{2mm}}c@{\hspace{2mm}}c@{\hspace{2mm}}c@{\hspace{2mm}}c@{\hspace{0mm}}}
   \includegraphics[width=\ww,height=\hh]{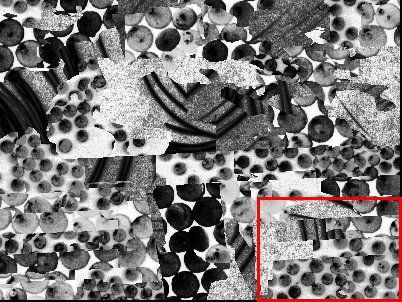}& 
  \includegraphics[width=\ww,height=\hh]{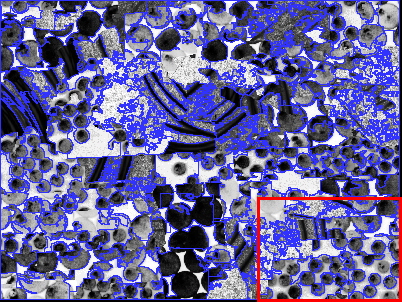}&
  \includegraphics[width=\ww,height=\hh]{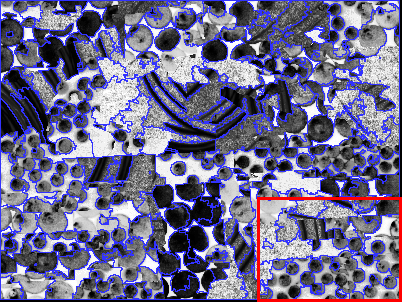}& 
  \includegraphics[width=\ww,height=\hh]{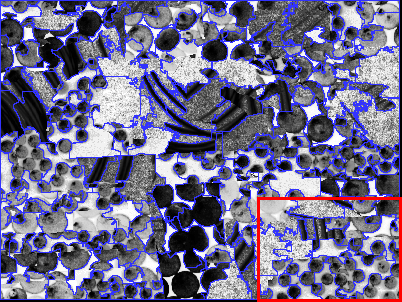}& 
  \includegraphics[width=\ww,height=\hh]{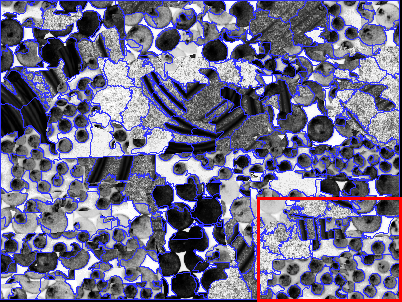}\\ %2015
  \includegraphics[width=\ww,height=\hh]{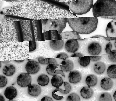}& 
  \includegraphics[width=\ww,height=\hh]{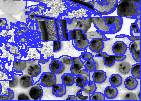}&
  \includegraphics[width=\ww,height=\hh]{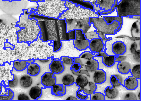}& 
  \includegraphics[width=\ww,height=\hh]{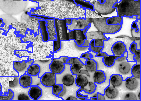}& 
  \includegraphics[width=\ww,height=\hh]{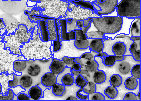}\\[0.75ex] %2015
  \includegraphics[width=\ww,height=\hh]{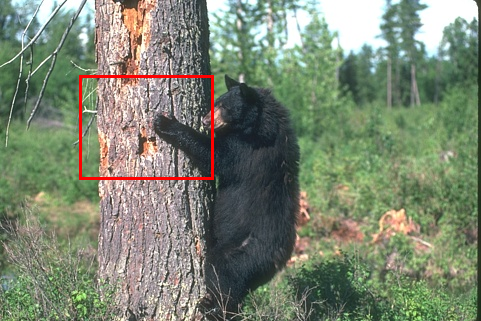}& 
  \includegraphics[width=\ww,height=\hh]{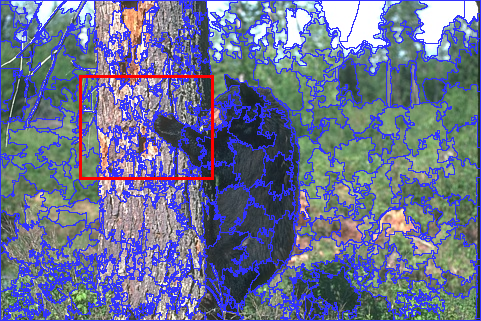}&
  \includegraphics[width=\ww,height=\hh]{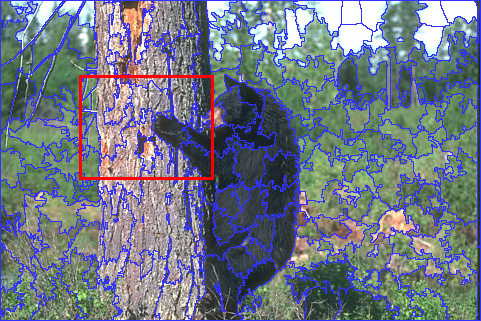}& 
  \includegraphics[width=\ww,height=\hh]{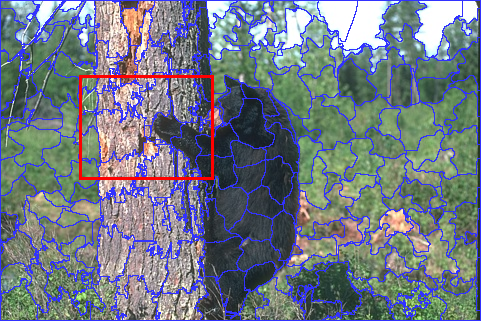}& 
  \includegraphics[width=\ww,height=\hh]{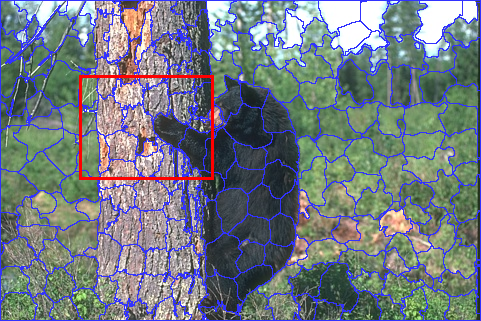}\\ %2015
  \includegraphics[width=\ww,height=\hh]{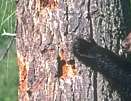}& 
  \includegraphics[width=\ww,height=\hh]{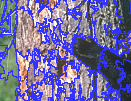}&
  \includegraphics[width=\ww,height=\hh]{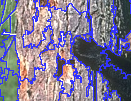}& 
  \includegraphics[width=\ww,height=\hh]{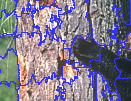}& 
  \includegraphics[width=\ww,height=\hh]{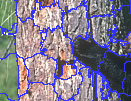}\\ %2015
  Initial image &LSC \cite{chen2017}  & SNIC \cite{achanta2017superpixels}  & SCALP \cite{giraud2018_scalp}&TASP  \\[-1ex]
 \end{tabular}
 }
 }
\caption{
Comparison between  TASP and most recent state-of-the-art methods.
TASP produces a more relevant result
on a natural texture composite image
(top)
and on a natural color image example from the BSD (bottom),  for $200$ superpixels.
} \vspace{-0.2cm}
\label{fig:comp_soa} 
\end{figure*}

\section{Results}
\label{sec:comp_soa}

\subsection{Validation Framework}

Similarly to \cite{randen1999filtering}, we create two new datasets
to evaluate texture segmentation performances. %
A highly challenging synthetic stripe (mix-Stripes) dataset of 10 images of size $300{\times}400$ pixels, is created by putting stripes similar to the ones in Figure \ref{fig:intro},
in variable shaped regions of minimum size $1000$ pixels.
Natural textures with normalized intensity are also taken from the Brodatz dataset \cite{brodatz1966},
to create 100 composite images (mix-Brodatz),
that can contain up to $10$ different textures.
Finally, we consider the standard Berkeley Segmentation Dataset (BSD) \cite{martin2001}, 
containing $200$ natural color  test images of size
$321{\times}481$ pixels.

Most parameters are empirically set once and for all, and their tuning has a  moderate impact on performances.
For the patch search, $N=8$ patches of size $5{\times}5$ pixels are selected outside a $\delta=3$ neighborhood.
In the clustering model, 
$m$ is set to $0.1$ and
$\beta$ to $25$  in \eqref{laser}. 
Color features $F$ are computed as in \cite{giraud2018_scalp}.
Finally,  the whole clustering process of
TASP  is performed in $10$ iterations as in \cite{achanta2012}.

 TASP is compared to the recent state-of-the-art methods
 SLIC \cite{achanta2012},
 ERGC \cite{buyssens2014},
 ETPS \cite{yao2015},
 LSC \cite{chen2017},
 SNIC \cite{achanta2017superpixels},
 and
 SCALP \cite{giraud2018_scalp}.
 Performances are evaluated with standard Achievable Segmentation Accuracy (ASA), and contour detection metric F-measure (F)
  as defined in 
\cite{giraud2017_gef}, and we report quantitative results
for an average number of $250$ superpixels.

\subsection{Influence of Contributions}

The visual impact of contributions is shown in Figure \ref{fig:unicity},
and ASA and F measures are reported in Table \ref{table:param} 
on the three considered datasets.
Our new texture homogeneity term \eqref{pbd} 
and constraint of texture unicity within superpixels \eqref{unicity}
both significantly improve performances on each data type. 
\vspace{-0.1cm}
\begin{table}[h]
\centering
\renewcommand{\arraystretch}{0.975}
{\color{black}
{\footnotesize
 \begin{tabular}{@{\hspace{0mm}}l@{\hspace{2mm}}c@{\hspace{1mm}}c@{\hspace{2mm}}c@{\hspace{1mm}}c@{\hspace{2mm}}c@{\hspace{1mm}}c@{\hspace{
 0mm}}}
  &  \multicolumn{2}{c}{mix-Stripes \text{ \hspace{0.1cm}}}&  \multicolumn{2}{c}{mix-Brodatz \text{ \hspace{0cm}}} & \multicolumn{2}{c}{BSD \text{ \hspace{-0.15cm}}} \\ \hline
  Method  & ASA & F & ASA & F & ASA & F \\\hline
  TASP w/o \eqref{pbd},\eqref{unicity}    & $0.8303$ & $0.3498$  & $0.7969$ & $0.4736$  & $0.9484$ & $0.4945$ \\
  TASP w/o \eqref{unicity}                & ${0.8486}$  & ${0.3882}$   & ${0.8112}$ & ${0.4812}$  &  ${0.9493}$   & ${0.4961}$ \\
  \textbf{TASP}                   & $\mathbf{0.8706}$     & $\mathbf{0.4232}$     & $\mathbf{0.8139}$ & $\mathbf{0.4824}$  & $\mathbf{0.9503}$  & $\mathbf{0.4992}$\\[-2.5ex]
 \end{tabular}
 }
 }
 \caption{Influence of contributions in the TASP method. 
} \vspace{-0.3cm}
 \label{table:param}
\end{table}

\subsection{Comparison to the State-of-the-Art Methods}

TASP is compared to state-of-the-art  approaches 
on an image similar to the mix-Stripes dataset in Figure \ref{fig:intro},
and to the most recent methods on mix-Brodatz and BSD images in Figure \ref{fig:comp_soa}.

A quantitative evaluation is also performed on the three datasets 
in Table \ref{tab:ressoa}.
TASP significantly increases the performances on the synthetic (mix-Stripes) and 
natural (mix-Brodatz) texture datasets, 
demonstrating its ability to
provide texture-aware superpixels.
TASP also obtains the best results on natural color images (BSD), 
using the same parameters,
while 
other methods fail at providing 
accurate results on the three data types at the same time.
Note that compared methods are used with parameters recommended by the authors.
Nevertheless, no other approach explicitly
captures texture information, so
TASP with default parameters still outperform state-of-the-art methods manually optimized for each dataset.

\vspace{-0.15cm}

\begin{table}[h] 
\centering
{\color{black}
{\footnotesize
 \begin{tabular}{@{\hspace{0mm}}l@{\hspace{2mm}}c@{\hspace{1mm}}c@{\hspace{3mm}}c@{\hspace{1mm}}c@{\hspace{3mm}}c@{\hspace{1mm}}c@{\hspace{
 0mm}}}
  &  \multicolumn{2}{c}{mix-Stripes \text{ \hspace{0.1cm}}}&  \multicolumn{2}{c}{mix-Brodatz \text{ \hspace{0cm}}} & \multicolumn{2}{c}{BSD \text{ \hspace{-0.15cm}}} \\ \hline
  Method & ASA & F & ASA & F & ASA & F\\\hline
   SLIC \cite{achanta2012}              & $0.7256$  & $\underline{0.4048}$  & $0.7784$ & $0.4607$                   & $0.9445$ & $0.4706$ \\ 
   ERGC \cite{buyssens2014}              & $0.6107$  & $0.3717$             & $0.7796$ & $0.4677$                    & $0.9477$ & $0.4571$\\ 
   ETPS \cite{yao2015}                    & $\underline{0.7769}$& $0.2953$  & $0.7568$ & $0.4354$                    & $0.9433$ & $0.4710$\\ 
   LSC \cite{chen2017}                    & $0.6979$ & $0.3156$             & $0.7908$ & $0.4552$                   & $\underline{0.9503}$ & $0.4421$ \\ 
   SNIC \cite{achanta2017superpixels}    & $0.6659$ & $0.3529$              & $0.7662$ & $\underline{0.4815}$        & $0.9410$ & $0.4617$ \\ 
   SCALP \cite{giraud2018_scalp}         & $0.7307$  & $0.3290$              & $\underline{0.7977}$& $0.4759$       & $0.9499$ & $\underline{0.4914}$   \\ 
  \textbf{TASP}                           & $\mathbf{0.8706}$  & $\mathbf{0.4232}$  
                                          & $\mathbf{0.8139}$& $\mathbf{0.4824}$ 
                                         & $\mathbf{0.9503}$ & $\mathbf{0.4992}$ \\[-1.25ex]  
 \end{tabular}
 }
 }
 \caption{TASP compared to the state-of-the-art methods.
Best and second results are respectively bold and underlined.} 
\vspace{-0.3cm}
 \label{tab:ressoa}
\end{table}

  \section{Conclusion}

  In this paper, 
we address the severe non-robustness of superpixel approaches to texture images by proposing a texture-aware decomposition method.
A new patch-based framework is introduced
to gather pixels having both similar color and local textural properties.
The proposed method is general, 
removes the need for manual setting of regularity constraint,
and also naturally can extend to generate supervoxels.

We outperform the segmentation of state-of-the-art methods on color, and synthetic and natural texture datasets, by using the same parameters.
This work
opens the way for larger use of superpixels and 
efficient application
of our approach
to medical image segmentation or video object tracking.

\newpage

{
\bibliographystyle{ieeetr}
\bibliography{biblio}
}

\end{document}